# A Prior Embedding-Driven Architecture for Long Distance Blind Iris Recognition

Qi Xiong[1,2]

1, International College, Hunan University of Arts and Sciences, Changde 415000, China

Xinman Zhang[2,*]

2, School of Automation Science and Engineering, Faculty of Electronic and Information Engineering; MOE Key Lab for Intelligent Networks and Network Security, Xi'an Jiaotong University, Xi'an 710049, Shaanxi, China

Jun Shen[3]

3, School of Computing and Information Technology, University of Wollongong, Wollongong, NSW 2522, Australia

---

Abstract: Blind iris images, which result from unknown degradation during the process of iris recognition at long distances, often lead to decreased iris recognition rates. Currently, little existing literature offers a solution to this problem. In response, we propose a prior embedding-driven architecture for long distance blind iris recognition. We first proposed a blind iris image restoration network called Iris-PPRGAN. To effectively restore the texture of the blind iris, Iris-PPRGAN includes a Generative Adversarial Network (GAN) used as a Prior Decoder, and a DNN used as the encoder. To extract iris features more efficiently, we then proposed a robust iris classifier by modifying the bottleneck module of InsightFace, which called Insight-Iris. A low-quality blind iris image is first restored by Iris-PPRGAN, then the restored iris image undergoes recognition via Insight-Iris. Experimental results on the public CASIA-Iris-distance dataset demonstrate that our proposed method significantly superior results to state-of-the-art blind iris restoration methods both quantitatively and qualitatively, Specifically, the recognition rate for long-distance blind iris images reaches 90% after processing with our methods, representing an improvement of approximately ten percentage points compared to images without restoration.

---

This research was funded by Hunan University of Arts and Science Research Project（NO. 23ZZ07）

*Corresponding author: Xinman Zhang

## 1、Introduction：

As biometric technology becomes increasingly prevalent in security authentication, iris recognition has gained considerable attention for its unique features and resistance to forgery [1,2]. However, in certain challenging environments, such as when iris images are captured from a distance in uncontrolled settings, various factors—such as low resolution, blur, and noise, or combinations of these issues—often cause significant degradation in image quality, as shown in Figure 1.

From Figure 1(a), it is evident that the iris image captured at a distance of three meters using specialized equipment has very clear textures, making it suitable for iris classification and recognition. However, in Figure 1(b), due to the influence of certain environmental factors, the captured iris image is very blurry, making the iris textures difficult to recognition. This situation will severely affect the accuracy of iris recognition [3,4].

Current hardware devices are often constrained by cost and technological limitations, making it challenging to achieve satisfactory recognition results for degraded iris images. Consequently, enhancing algorithms to improve iris recognition rates has become a prominent research focus. Among these enhancement methods, iris image restoration algorithms are particularly noteworthy. Their objective is to convert low-quality, low-resolution iris images into high-quality, high-resolution ones, thereby increasing the robustness of iris recognition systems.

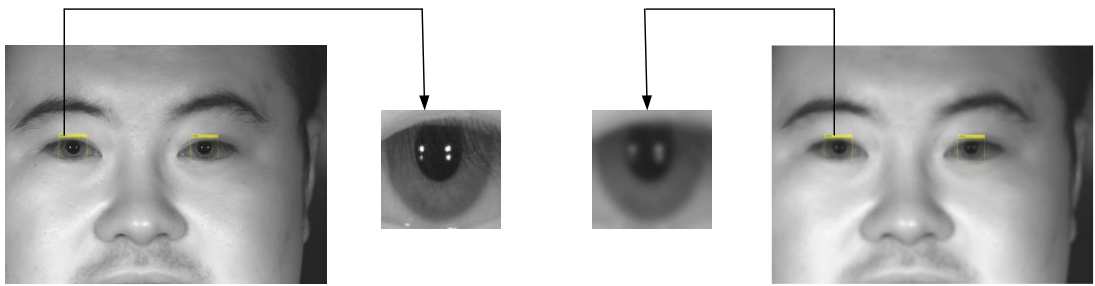

Figure 1(a) High-quality iris image          Figure 1(b) Degraded iris image

Figure 1: Comparison of High-Quality and Degraded Iris Images

In recent years, deep neural networks (DNNs) have shown exceptional performance across various computer vision tasks [5-8]. Numerous image restoration methods based on DNNs have been developed, demonstrating superior performance compared to traditional techniques [9-11].

The SRCNN algorithm utilized convolutional neural networks (CNNs) for image super-resolution, learning the mapping relationship between low-resolution and high-resolution images directly through these networks, thus achieving end-to-end super-resolution reconstruction [12,13]. This algorithm is noted for its simple structure and high training efficiency. However, due to the CNN's inherent limitation of local perception, it struggles to recover fine image details, resulting in suboptimal reconstruction quality.

With the rapid advancement of Generative Adversarial Networks (GANs) [14,15]. A several methods have recently emerged for reconstructing iris images from low-resolution inputs. The SRGAN algorithm utilizes the GAN framework and introduces a perceptual loss function that accounts for human visual perception [16]. This approach not only excels in PSNR metrics but also produces visual effects that align with human perception, effectively restoring image details. Building on this, ESRGAN replaces the simple residual blocks in SRGAN with Residual Dense Blocks (RDB). This modification allows ESRGAN to achieve superior visual effects in image super-resolution tasks, particularly in restoring image details and textures, demonstrating significant improvements over SRGAN [17].

The aforementioned methods have varying degrees of success in enhancing iris image quality and recognition rates. However, these studies typically use predetermined methods to generate degraded iris images. In real-world situations, the specific causes of iris image degradation are often unknown. As a result, many researchers have introduced the concept of blind image restoration.

Blind Image Restoration, or Blind Image Repair, involves restoring damaged images without prior knowledge of the cause or extent of the damage [18]. Due to the unique characteristics and position of the iris, blind iris image restoration remains a challenging research problem.

This paper's research approach employs a GAN to enhance low-quality blind iris images, transforming them into high-quality iris images. Once the image quality is restored, the high-quality iris images are input into a pre-trained iris classifier for recognition. This classifier uses a CNN structure, which excels at accurately extracting and classifying iris features. This end-to-end design not only enhances the usability of low-quality iris images but also improves the overall robustness and accuracy of the iris recognition system.

In this paper, we leverage the strengths of CNNs and GANs to propose a novel solution for recognizing long-distance, low-quality blind iris images. The main contributions of this paper are as follows:

(1) We proposed a network named Iris-PPRGAN for blind iris image restoration. Specifically, Iris-PPRGAN incorporates a GAN-based prior network as the decoder and a DNN as the encoder to effectively restore the texture of the blind iris.

(2). To effectively extract iris features and further enhance the robustness of iris recognition, we propose an iris classifier named Insight-Iris. This network is used not only for iris recognition tasks but also in Iris-PPRGAN to ensure that the generated iris remains consistent with the original iris in terms of identity characteristics.

(3) Our method sets new state-of-the-art in blind iris recognition. It has the capability to tackle low-quality iris images caused by various factors.

## 2 Related Works

2.1 **Iris recognition.** Iris recognition technology distinguishes individuals by using algorithms to model the unique textures of their irises. Daugman was the first to propose a successful commercial iris recognition system [19]. This system uses calculus operators to detect the inner and outer boundaries of the iris, and 2D Gabor filters to extract iris features. Recognition is then performed by calculating the Hamming distance. While Daugman's method shows excellent performance with ideal iris images, its recognition rate diminishes significantly with non-ideal images. Since then, many researchers have proposed improved methods [20-22]. However, these enhancements are still primarily based on ideal iris images, limiting their effectiveness in real-world scenarios.

In recent years, the rapid development of deep learning technology has led to significant progress in iris recognition. Deep learning models can automatically extract features, enhancing the accuracy and robustness of iris recognition systems. Nguyen et al. [23] were among the first to investigate the performance of pre-trained CNNs in iris recognition. They discovered that although these CNNs were initially trained to classify general objects, they were also effective in representing iris images. This approach successfully extracted discriminative visual features and achieved satisfactory recognition results on two iris datasets. Minaee et al. [24] explored the application of deep features extracted from VGG-Net for iris recognition. Their method was tested on two well-known iris databases and demonstrated satisfactory results. Luo et al. [25] designed a deep learning model incorporating spatial attention and channel attention mechanisms. These mechanisms were directly integrated into the feature extraction module, enabling the model to efficiently learn the most important features while suppressing unnecessary ones. Hafner et al. [26] adapted the Daugman-defined iris recognition pipeline by using the DenseNet-201 convolutional neural network as the feature extractor. This adaptation achieved a recognition accuracy of 97.3%.

The iris features extracted through deep learning are often utilized in iris classification networks. Gangwar et al. [27] proposed two backbone networks for iris recognition: DeepIrisNet-A and DeepIrisNet-B. DeepIrisNet-A is based on standard convolutional layers, while DeepIrisNet-B incorporates multiple Inception layers within its neural network architecture. Experimental results indicate that both networks are effective and robust in iris recognition tasks.

Given the similarities between iris recognition and general image classification tasks, various backbone network architectures have been adopted for iris recognition. For example, Zhao et al. utilized a capsule network based on the InceptionV3 architecture [28], while Hsiao et al. [29] employed EfficientNet as their backbone network.

**2.2 Iris Image Restoration.** We reviewed all literature on iris image enhancement from the past five years. The techniques discussed in these papers can be categorized into three types: traditional algorithms, CNN-based methods, and GAN-based methods. Traditional algorithms encompass methods that do not utilize deep learning. For instance, Liu M et al. [30] used a fuzzy filter on the region outside the iris boundary to reduce interference, allowing deep learning models to better focus on the iris features. While this method improved the recognition rate, it did not fundamentally enhance low-quality iris images. Additionally, papers [4] and [31] enhanced the CLAHE algorithm from different perspectives, essentially using interpolation methods.

Papers [12] and [32] explore the use of CNN to achieve super-resolution in iris images. Paper [12] focuses on evaluating different CNN architectures, highlighting the importance of maintaining texture details to generate more realistic images. In contrast, Paper [32] investigates various CNN architectures and also examines how image reprojection can enhance the accuracy of iris recognition systems. Furthermore, Paper [13] introduces an efficient iris image super-resolution network (ESISR). This network significantly reduces computational costs by minimizing the number of parameters and employing a sharpness-based loss function, all while maintaining image quality. This makes ESISR particularly suitable for mobile device applications.

Paper [33] proposes a method that utilizes a densely connected convolutional network as the generator. This approach combines adversarial learning with an identification loss function for joint training, thereby enhancing both the super-resolution quality and the recognition capability of iris images. Building on this, Paper [34] introduces the ocular super-resolution network (OSRCycleGAN), which is based on a cycle-consistent generative adversarial network (CycleGAN). This method aims to achieve super-resolution reconstruction from low-resolution ocular images.

Paper [35] introduces a super-resolution method based on GANs known as DDA-SRGAN. This method employs a dual-dimensional attention mechanism to automatically identify important regions of interest (ROIs) in the image and enhance the detailed features of these regions.

**2.3 Blind Image Restoration Techniques**

The aforementioned literature primarily addresses non-blind image restoration. Currently, there is no research specifically focused on blind iris restoration. However, several techniques have been developed for blind face restoration, including HiFaceGAN [36], PSFR-GAN (Progressive Semantic-Aware Style Transformation for Blind Face Restoration) [37], PULSE (Photo Upsampling via Latent Space Exploration) [38], and GPEN (Generative Facial Prior-Embedded Network) [18], among others.

HiFaceGAN is a GAN specifically designed for face image restoration. Its generator includes multiple sub-networks, each responsible for different levels of restoration tasks. PSFR-GAN progressively restores face images by utilizing semantic information and style transformation to improve the quality of restoration. PULSE is a super-resolution technique that does not directly optimize pixel-level losses. Instead, it finds the high-resolution output that best matches the low-resolution input through latent space exploration. GPEN is a method for face restoration and enhancement that employs a generative network embedded with facial prior knowledge. This approach effectively addresses various issues in face images, such as blurriness, noise, and compression artifacts. This paper can draw on the above-mentioned blind face restoration techniques to achieve blind iris restoration.

This paper aims to adapt the aforementioned Iris recognition technology and blind face restoration techniques to achieve blind iris recognition. We use a GAN as a Prior Decoder and a Pre-Trained Insight-Iris model to construct a new blind iris image restoration network. Insight-Iris is a robust iris classifier, which we developed by modifying the bottleneck module of InsightFace.

**3. Proposed Method**

**3.1 System Framework**

Given the unknown reasons for the degradation of blind iris images, this paper proposes to integrate a pre-trained GAN generator with an iris classifier into a single network. By freezing certain parts of the GAN network and fine-tuning the iris classifier, the method aims to achieve end-to-end recognition of low-quality iris images. This approach simplifies the process and has the potential to enhance the overall performance of the iris recognition system. The framework of the system is illustrated in Figure 2.

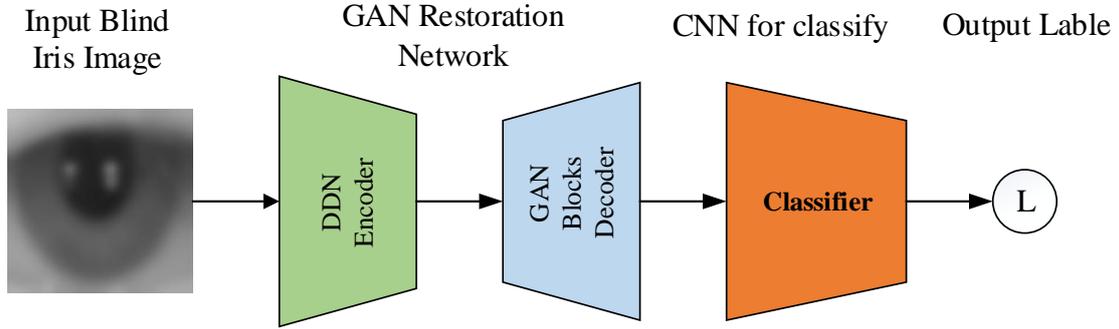

Figure 2 Overall System Framework

**3.2 Implementation of Iris Image Restoration Network**

Figure 2 clearly shows that the performance of the GAN network is pivotal to the recognition rate when restoring images. The restoration of blind iris images discussed in this chapter exemplifies a classic ill-posed inverse problem. The objective of this paper is to derive a high-quality iris image, denoted as $y$, from a low-quality blind iris image, denoted as x, as illustrated in Equation (1).

$$x = D(y) \qquad (1)$$

where D represents the degradation function (i.e., blurring, adding noise, etc.).

The primary challenge in solving the blind image restoration problem is the non-uniqueness of the solution. Many different high-quality images (y') can satisfy (x = D(y')), meaning that various high-quality iris images (y') might degrade into the same low-quality iris image (x). This is especially true when the degradation process results in the loss of critical information.

Existing methods [12,13,32] generally employ pixel-level loss functions to train DNNs for mapping $x$ to y. As a result, the final output tends to be an average of all high-quality iris images, often lacking detailed features and textures.

**3.2.1 Overall Network Structure**

To address the lack of detail and texture in iris images generated by the GAN network, this paper first trains a GAN prior network using StyleGAN technology, as shown in Figure 3. This GAN prior network is then embedded into a DNN as the decoder for high-quality iris image restoration, with the DNN serving as the encoder, as illustrated in Figure 4.

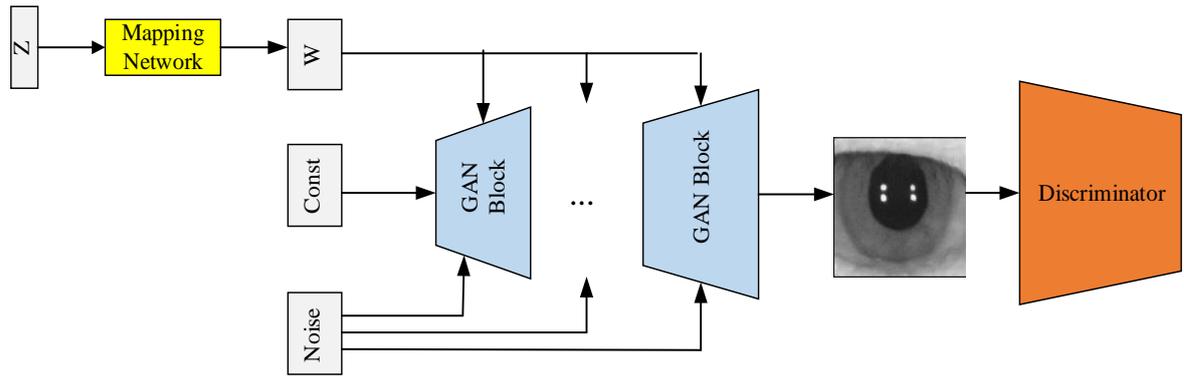

Figure 3: GAN prior network

Figure 4 presents the overall framework of the iris image restoration network, which is divided into three main components. The first component is a U-shaped generator that includes an encoder and a decoder. The second component is the discriminator. The third component is responsible for loss calculation.

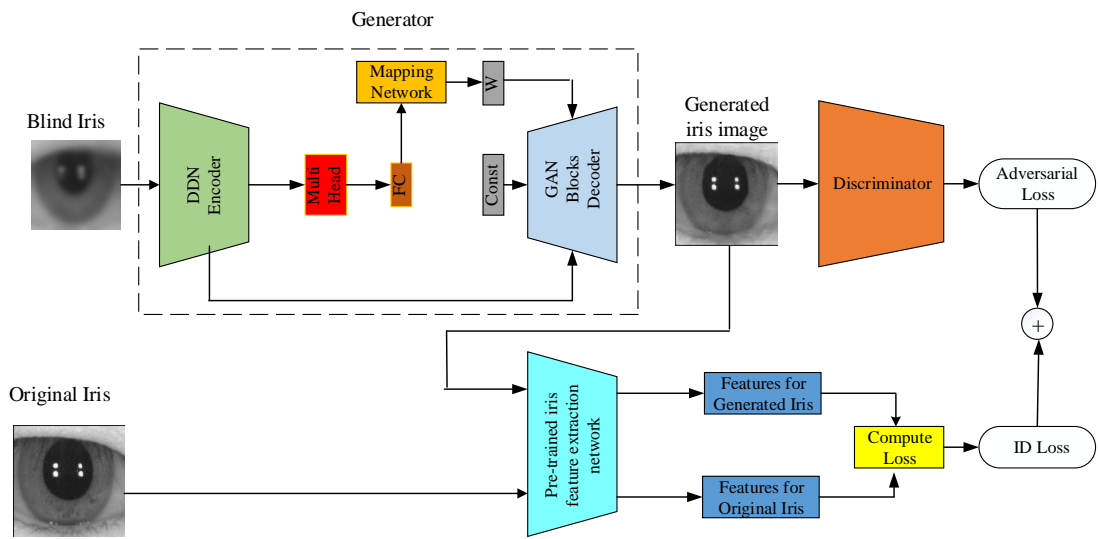

Figure 4 Overall framework of GAN network

The low-quality blind iris image, captured from a distance, is first processed by a DNN encoder equipped with a multi-head attention mechanism. This encoder's task is to convert the input iris image into a latent code ($z$), as illustrated in Figure 3. This latent code resides in the GAN's latent space ($Z$), a high-dimensional space that represents the intrinsic features of the data. Through this mapping, the encoder captures the critical information from the low-quality iris image and converts it into a more abstract representation, which can subsequently be used to generate or reconstruct the image.

The GAN prior network then reproduces the desired high-quality iris image $y$ through G($z$), where $G$ is the GAN generator trained during the learning phase. This generation process performs a one-to-one mapping, greatly reducing the uncertainty inherent in the one-to-many mappings of previous methods. Although GAN inversion methods [33-35] share similar concepts with the approach proposed here, they depend on pre-trained GAN models without further adjustments, leading to inconsistent results when processing iris images. In contrast, the method proposed in this section involves carefully designing, pre-training, and fine-tuning the GAN module to effectively restore blind iris images.

To ensure that the processed images retain the individual's unique features and recognition attributes, the pre-trained iris classifier is used as a feature extraction network, as shown in Figure 4. This classifier calculates the identity loss between the generated image and the original high-quality iris image. The GAN model presented in Figure 4 is named Iris-PPRGAN, which will be explained in detail in the subsequent sections.

### 3.2.2 GAN-Based Prior Network

U-Net [39] has been successfully applied to various image restoration tasks and is effective in preserving image details. Consequently, our Iris-PPRGAN adopts a U-shaped encoder-decoder architecture, as depicted in Figure 4. The GAN prior network is designed to meet two requirements: firstly, it must be capable of generating high-definition iris images; secondly, it should be easily integrated into the U-shaped network structure to function as a decoder.

Inspired by recent GAN architectures like StyleGAN [40,41], this paper employs a mapping network to project the latent code (z) into a less entangled space, as shown in Figure 3. The intermediate code (w) is then distributed to each GAN block. When integrating the GAN prior network into a DNN for fine-tuning, feature maps generated from each encoder layer are directly passed to corresponding decoder layers as noise inputs for the GAN blocks, as illustrated in Figure 4. This integration helps restore image details and contextual information because these feature maps carry significant spatial information from the input data. Therefore, when embedding the GAN prior network into a U-shaped DNN, it is essential to reserve enough space for these feature maps, allowing for their effective integration and utilization. This enhances the overall network performance and fine-tuning effectiveness.

The structure of the GAN blocks follows the architecture of StyleGAN v2, known for its superior ability to generate high-quality images. The number of GAN blocks matches the number of convolutional layers in the U-shaped DNN that can extract skip features, which is determined by the resolution of the input iris images.

### 3.2.3 DNN Encoder with Multi-Head Attention Mechanism

Given that the input low-quality iris image has a size of 256×256×3, a DNN comprising seven convolutional layers is employed as the encoder. The encoder extracts key features from the low-quality image to generate the input controls for the GAN, including the latent code and noise inputs. The network structure of the encoder is detailed in Table 1.

Table 1 Structure of the Encoder Network

| Layer | Number of Filters | Kernel Size | Stride | Feature Map Size |
|---|---|---|---|---|
| Input Image | - | - | - | 256×256×3 |
| Conv Layer 1 | 128 | 1×1 | 1 | 256×256×128 |
| Conv Layer 2 | 256 | 3×3 | 2 | 128×128×256 |
| Conv Layer 3 | 512 | 3×3 | 2 | 64×64×512 |
| Conv Layer 4 | 512 | 3×3 | 2 | 32×32×512 |
| Conv Layer 5 | 512 | 3×3 | 2 | 16×16×512 |
| Conv Layer 6 | 512 | 3×3 | 2 | 8×8×512 |
| Conv Layer 7 | 512 | 3×3 | 2 | 4×4×512 |

The design concept of the encoder is as follows:

1) Use the output of the fully connected layers (i.e., deeper features) to replace the latent code $z$. These deep features encapsulate high-level abstract information extracted from the input image, which is used to govern the global structure of the generated iris image.

2) Use the shallow outputs of the encoder to replace noise inputs. These shallow features contain more local and detailed information, which manage local aspects in the generated image, such as the texture of the iris. These features significantly enhance the detail richness and realism of the generated image.

Due to the iris's unique position between the sclera and the pupil, a multi-head attention mechanism is added to the model's encoder to better focus on the iris region in the input image and ignore irrelevant areas such as the pupil and sclera. This module processes incoming feature maps to capture the interactive information of image features, emphasizing key information during feature transmission and effectively mitigating detail loss, as shown in Figure 5.

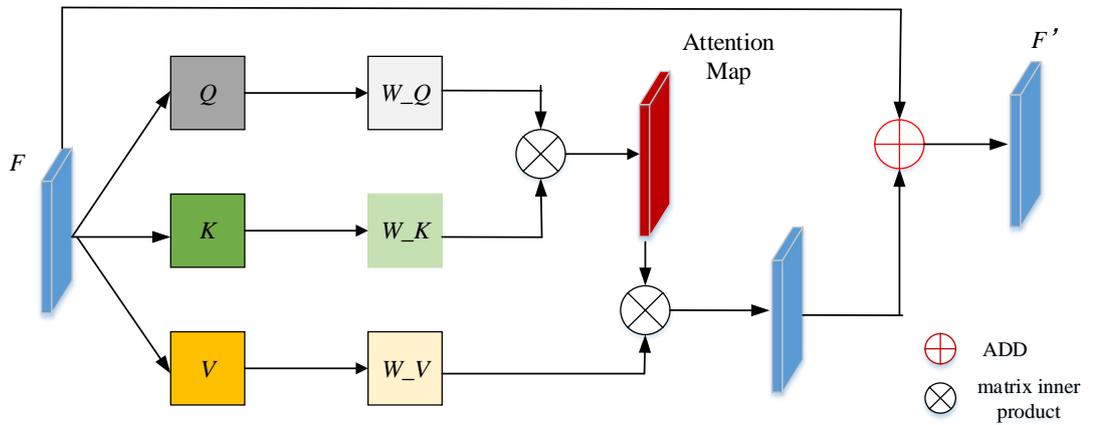

Figure 5 Multi-head attention module

The multi-head self-attention module used in this paper is based on CNNs. This module first defines the input dimensions and the number of heads, which are crucial for the multi-head self-attention mechanism. These parameters dictate how the attention mechanism is divided into multiple "heads" for parallel processing and determine the dimensional size each head processes. The input is split into multiple parts, or "heads," each of which independently performs the same attention operation. This design enables the model to capture different aspects of information in various representational subspaces, thereby enhancing its learning and generalization capabilities.

In the multi-head attention module shown in Figure 5, three dedicated convolution layers process the input data to produce the corresponding Q, K, and V representations. The output channel number of these convolution layers is the query dimension multiplied by the number of heads. Each convolution layer has a kernel size of 1 to maintain spatial dimensions while transforming feature dimensions.

During forward propagation, the input feature map (*F*) generates the query, key, and value feature maps through these three convolution layers, respectively. These feature maps are the core components of the attention mechanism and are rearranged to create independent feature subspaces for each head. The transposed product of the query and key is then calculated to obtain attention scores, which are normalized through a softmax layer to form the attention map. This attention map focuses on important information by performing matrix multiplication with the value feature map.

Finally, another convolution layer combines the multi-head outputs back to the original feature dimensions, while a residual connection adds the input back to the output to enhance the model's learning capability and gradient flow. This design allows the multi-head self-attention mechanism to improve the model's understanding of the input data while maintaining computational efficiency.

The implementation of this multi-head self-attention mechanism equips the model to handle complex image and sequence tasks effectively. It enables the model to learn richer and more discriminative feature representations across multiple subspaces, thus improving overall performance and efficiency.

After extensive experimental testing, incorporating the multi-head attention mechanism in the middle part of the network yielded the best results. In this configuration, the multi-head attention mechanism computes attention values on the downsampled feature maps, which are then weighted and integrated into these maps. The adjusted feature maps are subsequently passed to deeper network layers.

### 3.3.4 Discriminator Module and Loss Functions

The discriminator network used in the paper directly adopts the implementation of the StyleGAN2 discriminator. To fine-tune the proposed GAN model, we employ three loss functions: adversarial loss $L_A$, $L1$ smooth loss, and identity loss $L_{id}$. The adversarial loss $L_A$ is inherited from the GAN prior network.

$$L_A = \min_G \max_D E_{(X)} \log(1 + \exp(-D(G(\tilde{X})))) \qquad (2)$$

Where, X and $\tilde{X}$ represent the real high-definition image and the degraded low-quality image, respectively. *G* denotes the generator during training, and *D* denotes the discriminator.

Smooth $L1$ Loss, also known as Huber Loss, is commonly used in regression problems. It combines $L1$ and $L2$ loss to mitigate the impact of outliers on model training. Smooth $L1$ Loss uses the squared term for smaller errors and the absolute value for larger errors. This approach maintains the robustness of the loss function while reducing the risk of gradient explosion.

$$L1 = \begin{cases} 0.5 \cdot (X - G(\tilde{X}))^2 & \text{if } |X - G(\tilde{X})| < 1 \\ |X - G(\tilde{X})| - 0.5 & \text{otherwise} \end{cases} \quad (3)$$

where, X and $\tilde{X}$ represent the real high-definition image and the degraded low-quality image, respectively. G denotes the generator during training.

Identity loss $L_{id}$ relies on the dot product between feature vectors to measure similarity. The larger the dot product, the more similar the two are.

$$L_{id} = 1 - F(X) \bullet F(G(\tilde{X})) \quad (4)$$

Where, $F$ represents the pre-trained iris classifier used to extract feature vectors. The final loss $L$ is as follows:

$$L = L_A + L_1 + L_{id} \quad (5)$$

Smooth $L1$ Loss enhances fine image details and preserves original color information. Introducing identity loss $L_{id}$ helps balance the adversarial loss $L_A$, leading to the restoration of more realistic and identity-consistent iris images.

### 3.3 Implementation of the Classifier

As illustrated in Figure 2, the classifier is a pivotal component of our system. An effective classifier can better extract features from iris images, leading to higher recognition rates. Commonly used classifiers include VGG, ResNet50, and other deep CNNs. Recent advancements in deep learning have introduced even more powerful classifiers, such as the InsightFace framework used in face recognition [42].

In this paper, we introduce a classifier specifically designed for iris recognition, named Insight-Iris, which is based on the InsightFace framework. The structure of Insight-Iris is detailed in Figure 6.

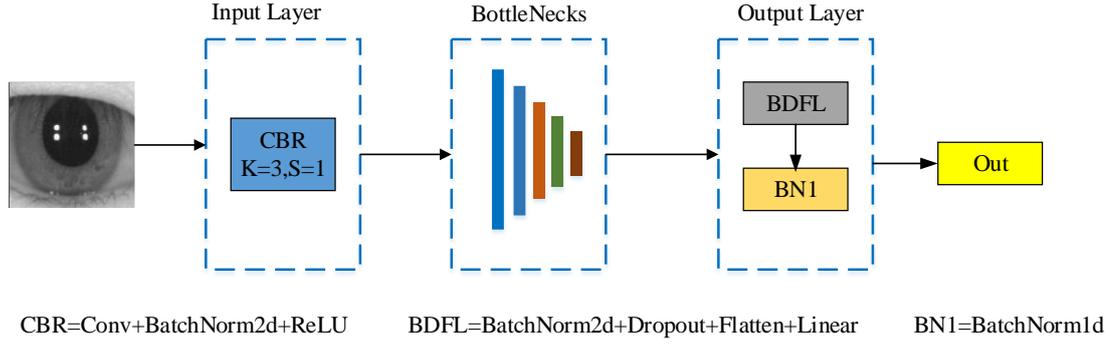

CBR=Conv+BatchNorm2d+ReLU     BDFL=BatchNorm2d+Dropout+Flatten+Linear     BN1=BatchNorm1d

(a) Overall Framework of Insight-Iris

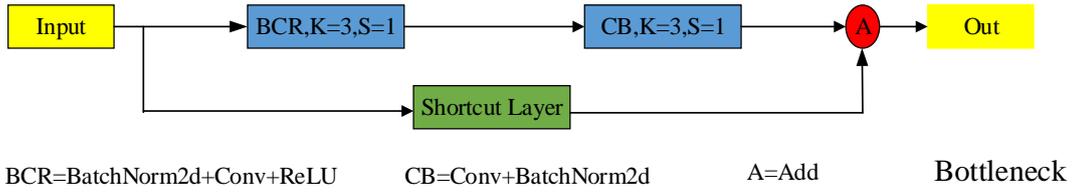

BCR=BatchNorm2d+Conv+ReLU     CB=Conv+BatchNorm2d     A=Add     Bottleneck

(b)Bottleneck structure. K represents the size of the convolutional kernel. S represents the stride

Figure 6: Network Structure of Insight-Iris

In Figure 6(a), the Insight-Iris consists of an input layer, bottleneck layers, and an output layer. The input layer is composed of a convolutional layer, a batch normalization layer, and a ReLU activation function. The model's main body comprises 50 bottleneck modules[43]. Each bottleneck module, as detailed in Figure 6(b), comprises two main components: the shortcut layer and the residual layer.

The shortcut layer employs a MaxPool2d layer with a kernel size of 1 for down-sampling. This step is followed by a Conv2d layer that uses a $1 \times 1$ kernel, coupled with a BatchNorm2d layer to align the dimensions. The residual layer is composed of several sequential steps. Initially, a BatchNorm2d layer standardizes the input. This is followed by a Conv2d layer with a $3 \times 3$ kernel that maintains the same input and output spatial dimensions. After this convolution, a BatchNorm2d layer and ReLU activation function further normalize and activate the features. Finally, another Conv2d layer with a $3 \times 3$ kernel adjusts the spatial dimensions, and a concluding BatchNorm2d layer once again normalizes the output. This layered structure ensures effective down-sampling, normalization, and activation, necessary for robust Blind Iris Restoration.

During forward propagation, the input data flows through both the shortcut and residual layers. Their outputs are then combined to produce the final output. This approach utilizes residual learning to improve the network's ability to extract features across multiple layers, thereby enhancing its representation capabilities.

The network's output layer first utilizes batch normalization and dropout operations to process the output from the convolutional layers. This enhances model stability and reduces overfitting. The multi-dimensional data is then flattened into a one-dimensional vector and passed through a fully connected layer to map the features to classification labels. To further stabilize the output, 1D batch normalization is applied again. This sequence effectively transforms deep features into the final classification result. Such a structure improves the network's training efficiency and generalization performance.

## 4 Experimental Results and Analysis

### 4.1 Experimental Dataset

In this experiment, we used the CASIA-Iris-Distance dataset from the Chinese Academy of Sciences . This dataset was created with an advanced biometric sensor capable of detecting iris and facial features up to 3 meters away. The high-resolution images include both irises and facial features and comprise 142 subjects with a total of 2,567 images. Each image has a resolution of 2352×1728 pixels. Figure 7 shows three example images from the CASIA-Iris-Distance database.

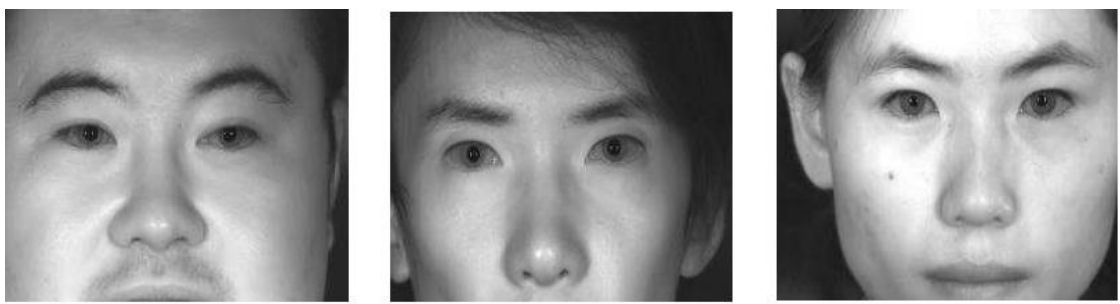

Figure 7. Example Images from the CASIA-Iris-Distance Database

From this dataset, we selected 141 classes of images. Of these, 102 classes containing a total of 1,748 left-eye iris images were used to train the GAN's prior network. Corresponding low-quality iris images were then synthesized from these high-quality images to fine-tune the Iris-PPRGAN network. To evaluate the model's performance, 80% of the remaining 39 classes, totaling 624 images, were used to train the iris classifier. The remaining 20%, or 156 images, served as the test set.

The low-quality iris images in the test set were degraded according to a blind image quality degradation principle, then processed by Iris-PPRGAN, and finally classified using the Insight-Iris classifier. To quantitatively assess the method, we used Peak Signal-to-Noise Ratio (PSNR), Frechet Inception Distances (FID) [18], and classification accuracy as performance metrics.

Importantly, there is no overlap between the dataset used to train the GAN network and the test set used to evaluate the network's performance. This ensures that the training and test datasets remain completely separate.

### 4.2 Training Strategy

The training of the entire network is organized into four steps:

1, Pre-train the GAN Prior Network: Select 102 classes of images from the experimental dataset and pre-train the GAN prior network following the training strategy of StyleGAN.

2, Train the Iris Classifier: Select an additional 39 classes of images to train the Insight-Iris classifier.

3, Fine-tune the Iris-PPRGAN Network: Embed the pre-trained GAN model into the proposed Iris-PPRGAN network and fine-tune the entire network using a set of synthesized low-quality (LQ) and high-quality (HQ) iris image pairs (the image synthesis process is detailed in section 4.3).

4, Integrate and Fine-tune: Combine the trained Iris-PPRGAN network with the trained iris classifier to create a system for long-distance low-quality iris classification. Utilize the input low-quality iris images, freeze the Iris-PPRGAN network, and fine-tune the iris classifier.

This structured approach ensures a cohesive and efficient training process for the entire network.

### 4.3 Implementation of Low-Quality Iris Images

Since the pre-trained GAN prior network must be embedded into Iris-PPRGAN for fine-tuning, we need to construct low-quality (LQ) to high-quality (HQ) image pairs. To achieve this, we use the following degradation model to synthesize degraded iris images from the 102 classes of high-quality (HQ) iris images.

$$I^d = (I \otimes \mathbf{k}) \downarrow_s + \mathbf{n}_\sigma \qquad (6)$$

Where, $I$, $\mathbf{k}$, $\mathbf{n}_\sigma$, $I^d$ represent the input iris image, blur kernel parameter, Gaussian noise intensity, and degraded image, respectively. $\otimes, \downarrow_s$ are represented as 2D convolution and standard s times downsampler. In the experiments, for each image, the blur kernel k is randomly selected from a set of blur models. The specific degradation parameters are shown in Table 2.

Table 2 Ranges of Degradation Parameters

|  | Parameter Name | Value Range |
|---|---|---|
| Blur Kernel k | Blur Kernel Type | ['iso', 'aniso'] |
|  | Blur Kernel Type Selection Probability | [0.5, 0.5] |
|  | Blur Kernel Size | 41 |
|  | Gaussian Blur Kernel Standard Deviation Range | [0.1, 10] |
|  | Blur Kernel Type | [0.8, 8] |
|  | Blur Kernel Type Selection Probability | [0, 20] |

Using these degradation parameters, 1,748 facial images were degraded. Despite the degradation, these images can still be processed by the improved YOLO network to detect and segment the iris area, as shown in Figure 1(b). Examples of the degraded iris images alongside their original high-quality counterparts are presented in Figure 8.

Figure 8 illustrates that, unlike traditional image degradation methods, the degradation model based on Formula (6) simulates the complex degradation process of blind images in uncontrolled environments. The extent of quality degradation varies among images; some exhibit only slight quality reduction, as shown in Figure 8(c), while others show significant quality deterioration, as depicted in Figure 8(d).

### 4.4 Experimental Results of the GAN Network

Fine-tuning the model with numerous severely degraded images, such as those in Figure 8, enables the DDN encoder component of the Iris-PPRGAN to learn to generate appropriate latent codes and noise. These inputs are then fed into the GAN prior decoder network, which is simultaneously updated to effectively handle severely degraded iris images in real-world scenarios.

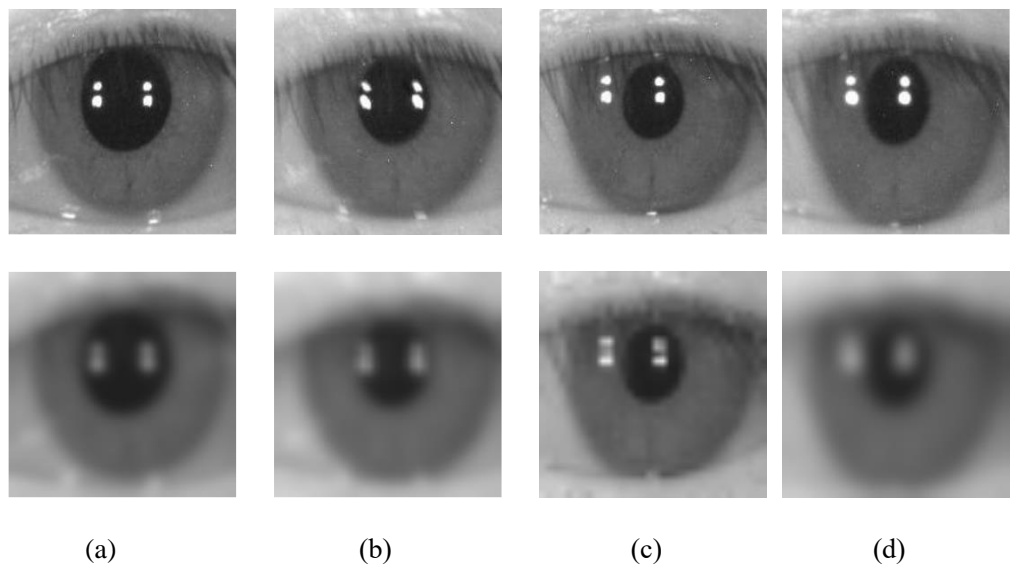

(a)　　　　　　(b)　　　　　　(c)　　　　　　(d)

Figure 8 Examples of Low-Quality to High-Quality Iris Image Pairs

During the model update process, we use the Adam optimizer with a batch size of 2. Different components of the GAN network—the encoder, decoder, and discriminator—are assigned different learning rates (LR). Specifically, the learning rate for the encoder is set to 0.0002. The learning rates for the decoder and discriminator are set to be 10 times and 100 times that of the encoder, respectively. During the testing phase, the discriminator is removed. Figure 9 compares the recovery effects of traditional image restoration algorithms and GAN models on the severely degraded image shown in Figure 8(d).

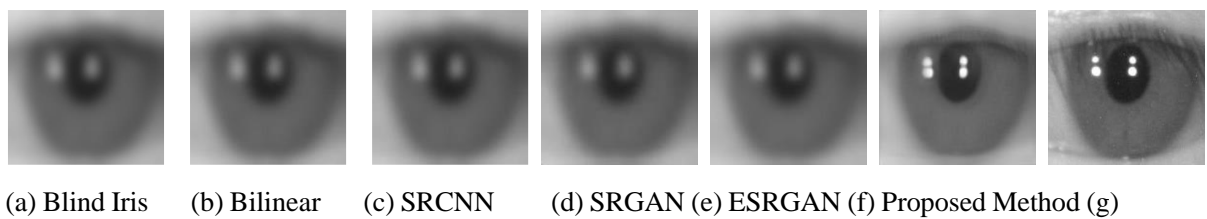

(a) Blind Iris　　(b) Bilinear　　(c) SRCNN　　(d) SRGAN (e) ESRGAN (f) Proposed Method (g) Original Image

Figure 9 Recovery Performance of Various GAN Models on Low-Quality Iris Images

Figures 9(b)-(e) demonstrate that while some GAN models perform well on images degraded in certain known ways, their performance on blind image restoration, as examined in this study, is inadequate. Figure 9(f) shows the recovery results of our proposed method. Although it does not fully restore the image to its original quality, it offers a significant visual improvement compared to the low-quality image in Figure

9(a). Moreover, for iris recognition, enhancing the appearance alone is insufficient; it is equally critical to improve the recognition rate.

### 4.5 Experimental Results of the Iris Classifier

The iris classifier in our method has a dual role: it enhances recognition accuracy and accurately extracts iris features to calculate the identity loss of the GAN network. Given that the input images are blind iris images similar to Figure 9(a), traditional methods that precisely locate the iris region are no longer feasible under such low-quality constraints. Therefore, we use the entire low-quality iris image as input. This approach not only simplifies computation but also enables the model to learn the global information of the iris image.

This study uses 80% of a total of 39 classes of iris images, amounting to 624 images, for the training set and employs transfer learning to train the iris classifier. The remaining 20%, or 156 iris images, are used as the test set. Table 3 presents the operating environments for the different deep convolutional neural networks, while Table 4 shows the classification recognition results. All experiments are conducted in a Windows 10 environment.

Table 3 Operating Environments for Various Classifiers

| Classifier | Python | PyTorch | Epoch | Batch | Learning Rating |
|---|---|---|---|---|---|
| VGG16 | 3.7.1 | 1.8.0 | 40 | 32 | $1\times10^{-3}$ |
| MobileNetv2 | 3.7.1 | 1.8.0 | 40 | 32 | $1\times10^{-3}$ |
| Resnet50 | 3.7.1 | 1.8.0 | 40 | 32 | $1\times10^{-3}$ |
| Vision Transformer(ViT) | 3.8.17 | 1.11.0 | 40 | 32 | $1\times10^{-3}$ |
| Insight-Iris | 3.8.17 | 1.11.0 | 20 | 16 | $1\times10^{-4}$ |

Table 4 Comparison of Recognition Rates for Various Classifiers

| Classifier | Recognition Rate on Original High-Definition Iris Images | Recognition Rate on Blind Iris Images |
|---|---|---|
| VGG16 | 99.46% | 33.33% |
| MobileNetv2 | 94.23 | 7.05% |
| Resnet50 | 95.51% | 14.10% |
| ViT | 98.08% | 35.26% |
| Insight-Iris | **98.74%** | **80.77%** |

As shown in Table 4, traditional deep convolutional networks achieve excellent recognition rates for high-quality iris images, with all exceeding 95%, and VGG16 nearing 100%. However, their recognition rates drop significantly for blind iris images, indicating poor robustness. In contrast, although Insight-Iris also experiences a decline, it still maintains a recognition rate of 80%, demonstrating strong robustness.

### 4.6 Recognition Experiment Results for Long-Distance Blind Iris

In the experiments, we use traditional image restoration algorithms and various GAN models to replace the GAN shown in Figure 2. Using the method described in section 4.2, low-quality iris images are generated from the 624 training images mentioned in section 4.1 and used as input to Figure 2 to fine-tune the Insight-Iris classifier. Table 5 presents the performance of the long-distance iris classifier under different image restoration algorithms used in Figure 2.

Table 5 Comparison of Recognition Rates for Low-Quality Iris Images Using Different Restoration Algorithms

| Restoration Algorithm | PSNR | FID | Recognition Rate |
|---|---|---|---|
| 无 | 16.54 | 50.39 | 80.77% |
| Bilinear | 16.39 | 49.40 | 80.12% |
| SRCNN | 16.37 | 49.09 | 80.12% |
| SRGAN | 15.52 | 49.15 | 78..20 |
| ESRGAN | 16.18 | 50.31 | 74.36% |
| Iris-PPRGAN | 15.57 | **45.43** | **90.38%** |

This table offers a clear comparison of the recognition rates for low-quality iris images when different restoration algorithms are applied. The quality of the restored images is evaluated using PSNR and FID metrics, and the resulting recognition rates highlight the effectiveness of each restoration method.

From Table 5, we observe that although the proposed method does not excel in PSNR performance, it achieves the lowest FID value. This suggests that the images generated by Iris-PPRGAN are more similar to the original images. Furthermore, a comparison between Tables 4 and 5 reveals that traditional interpolation algorithms and conventional deep learning-based image restoration methods result in lower recognition rates than the unprocessed blind iris images. In other words, these methods introduce artifacts that not only fail to improve recognition performance but actually degrade it. Although the proposed restoration model, Iris-PPRGAN, does not reach the recognition rate of the original high-definition images, it improves the recognition rate by nearly 10%, achieving a level of 90.38%.

### 4.7 Ablation Experiments

To investigate the effectiveness of the components of the proposed model, two types of ablation experiments were designed.

(1) Effect of Different Classifiers on Recognition Rate: Various classifiers were used to replace the iris classifier shown in Figure 2, in order to test the recognition rate of blind iris images. The results are documented in Table 6.

Table 6 Effect of Different Classifiers on Recognition Rate for Blind Iris Images

| Classifier | Recognition Rate on Blind Iris Images | Recognition Rate on Restored Iris Images |
| --- | --- | --- |
| VGG16 | 36.54% | 55.77% |
| MobileNetv2 | 5.13% | 26.28% |
| Resnet50 | 7.05% | 41.03% |
| ViT | 40.10% | 40.38% |
| Insight-Iris | **80.77%** | **90.38%** |

Table 6 shows that the recognition rate of blind iris images improves to varying degrees, regardless of which classifier is used in the system proposed in Figure 2. This demonstrates the strong scalability of the proposed system. However, the highest recognition rate is achieved only with the Insight-Iris network, highlighting the robustness of Insight-Iris.

(2) Effect of the Proposed System on Original Iris Images: Although the proposed system is designed for blind iris images, the quality of input images cannot be determined in practical applications. To verify the robustness of BIRN, we compared the performance of original, non-degraded iris images under different restoration algorithms. The experimental results are documented in Table 7.

Table 7 shows that the restoration algorithms have a slight impact on the recognition rate of original, non-degraded iris images. Unlike other restoration methods, our proposed Iris-PPRGAN not only maintains but actually improves the recognition rate of the original iris images by 0.62%. This demonstrates its strong robustness.

Table 7 Comparison of the Effect of Different Restoration Algorithms on the Original Iris Images

| Restoration Algorithm | Recognition Rate |
| --- | --- |
| Bilinear | 98.71% |
| SRCNN | 98.71% |
| SRGAN | 98.71% |
| ESRGAN | 98.07% |
| Iris-PPRGAN | **99.36%** |

**5 Conclusion**

This paper introduces a prior embedding-driven architecture for the recognition of long-distance blind iris images. The system integrates two key components: a novel iris image restoration network called Iris-PPRGAN, and a new iris classifier named Insight-Iris. In developing Iris-PPRGAN, we achieve high-quality iris image restoration by embedding a pre-trained GAN into a U-shaped deep neural network as a decoder, and fine-tuning the entire GAN network using low-quality iris images extracted from artificially degraded facial images. Subsequently, different low-quality iris images are fed into Iris-PPRGAN to fine-tune the iris classifier. Experimental results demonstrate that the methods we proposed can achieve a recognition rate of up to 90%, improving the recognition rate by 10% and outperforming existing techniques for low-quality iris restoration. This ensures robustness and accuracy in recognizing long-distance, low-quality blind iris images.

Acknowledgments: This research was funded by Hunan University of Arts and Science Research Project（NO. 23ZZ07）

# Referenes